\documentclass{article} 
\usepackage[table]{xcolor}
\usepackage[final]{colm2025_conference}
\usepackage{microtype}
\usepackage{hyperref}
\usepackage{url}
\usepackage{booktabs}
\usepackage{lineno}
\usepackage{graphicx}
\usepackage{lineno}
\usepackage{wrapfig}
\usepackage{multirow}
\usepackage{booktabs}
\usepackage{bbding}
\usepackage{graphicx}
\usepackage{adjustbox}
\usepackage{enumitem}
\usepackage{fontawesome5}

\newcommand{\DatasetName}{SciReplicate-Bench}
\newcommand{\ModelName}{Sci-Reproducer}

\definecolor{darkblue}{rgb}{0, 0, 0.5}
\hypersetup{colorlinks=true, citecolor=darkblue, linkcolor=darkblue, urlcolor=darkblue}

\newcommand{\thinking}{%
  \raisebox{-0.15\height}{\includegraphics[height=1em]{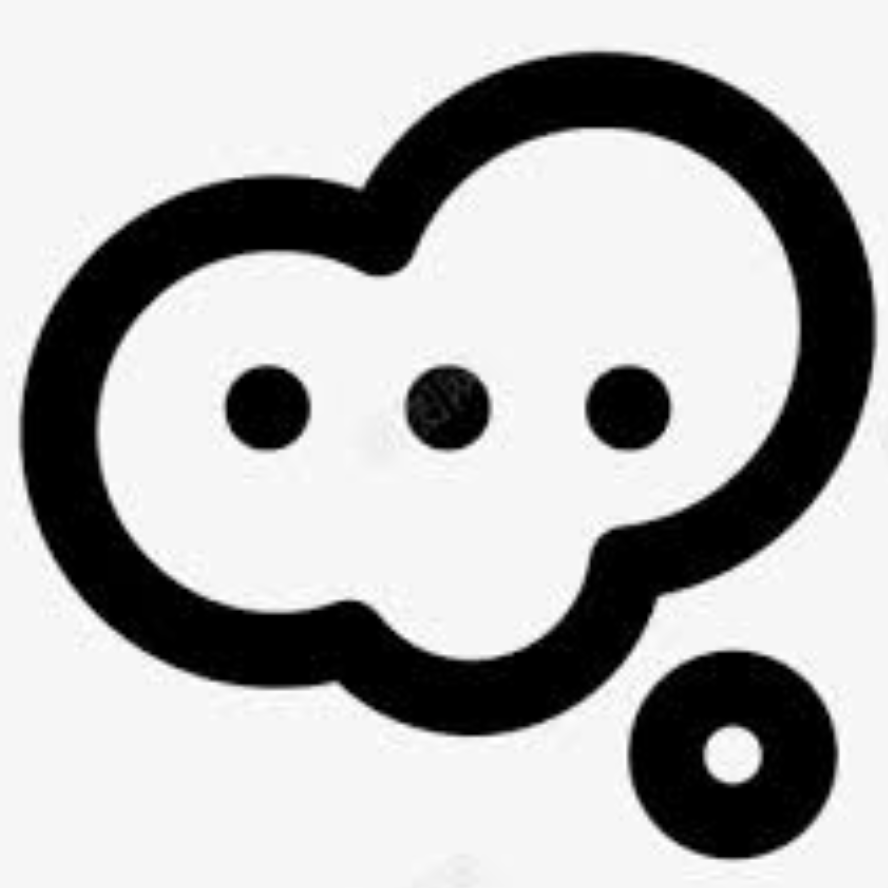}}
}

\newcommand{\gptmini}{%
  \raisebox{-.15\height}{\includegraphics[height=1.1em]{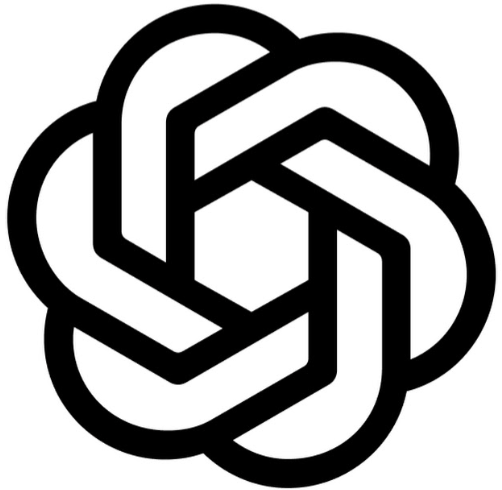}}~\textbf{GPT-4o-mini}%
}

\newcommand{\gpttwo}{%
  \raisebox{-.15\height}{\includegraphics[height=1.1em]{figures/openai.png}}~\textbf{GPT-4o}%
}

\newcommand{\olow}{%
  \raisebox{-.15\height}{\includegraphics[height=1.1em]{figures/openai.png}}~\textbf{o3-mini-low}%
}

\newcommand{\omedium}{%
  \raisebox{-.15\height}{\includegraphics[height=1.1em]{figures/openai.png}}~\textbf{o3-mini-medium}%
}

\newcommand{\ohigh}{%
  \raisebox{-.15\height}{\includegraphics[height=1.1em]{figures/openai.png}}~\textbf{o3-mini-high}%
}

\newcommand{\deepseek}{%
  \raisebox{-.15\height}{\includegraphics[height=1.1em]{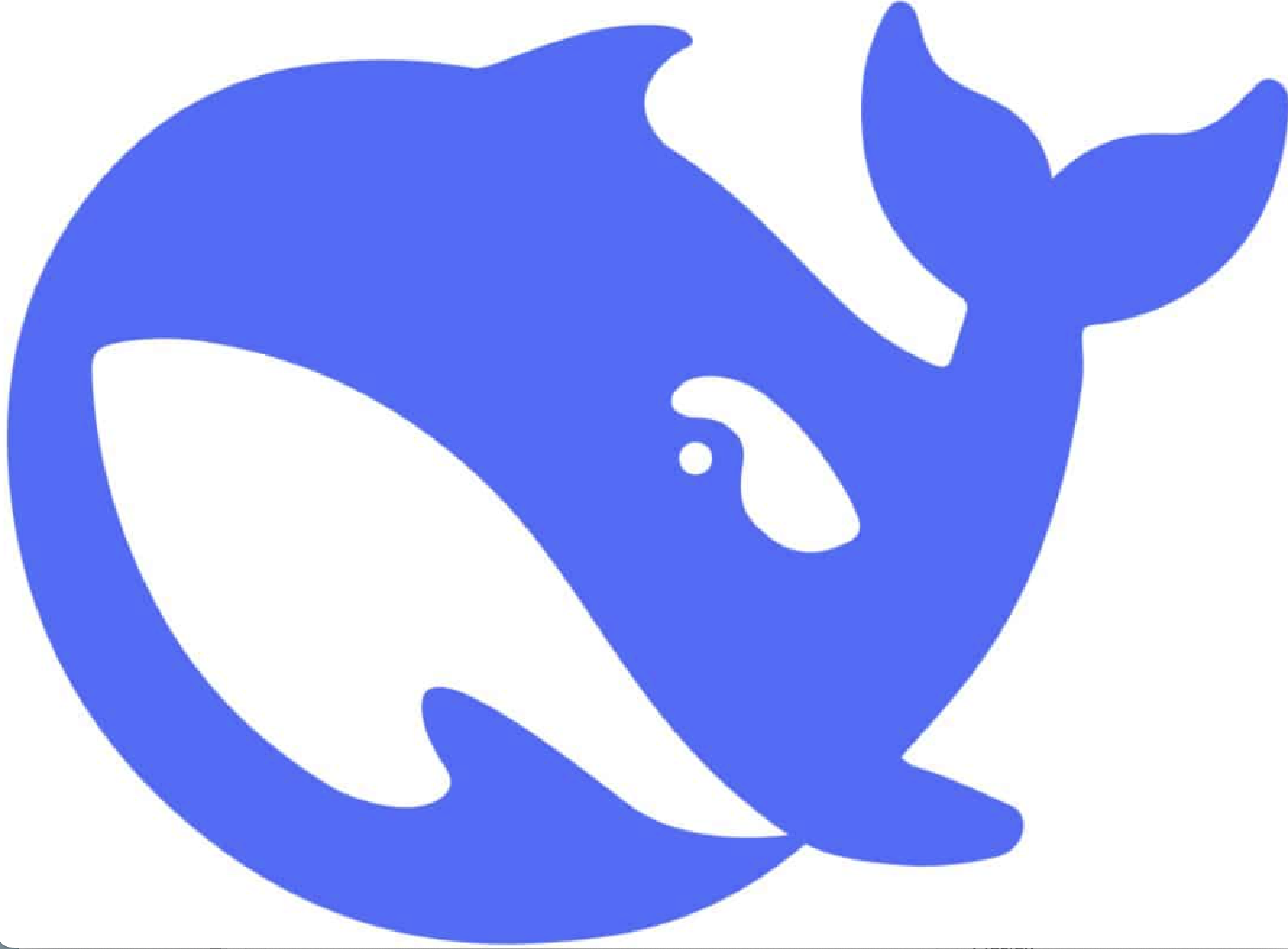}}~\textbf{Deepseek-V3}%
}

\newcommand{\gemini}{%
  \raisebox{-.15\height}{\includegraphics[height=1.1em]{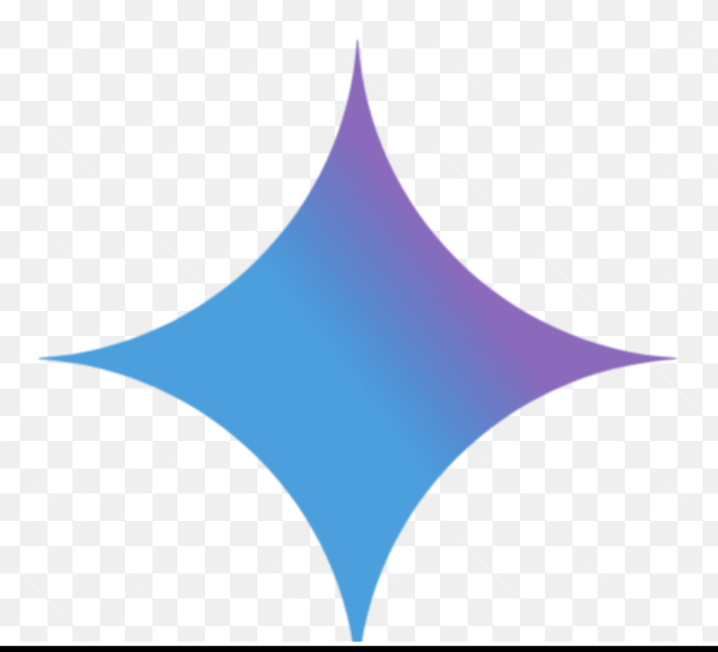}}~\textbf{Gemini-2.0-Flash}%
}

\newcommand{\claude}{%
  \raisebox{-.15\height}{\includegraphics[height=1.1em]{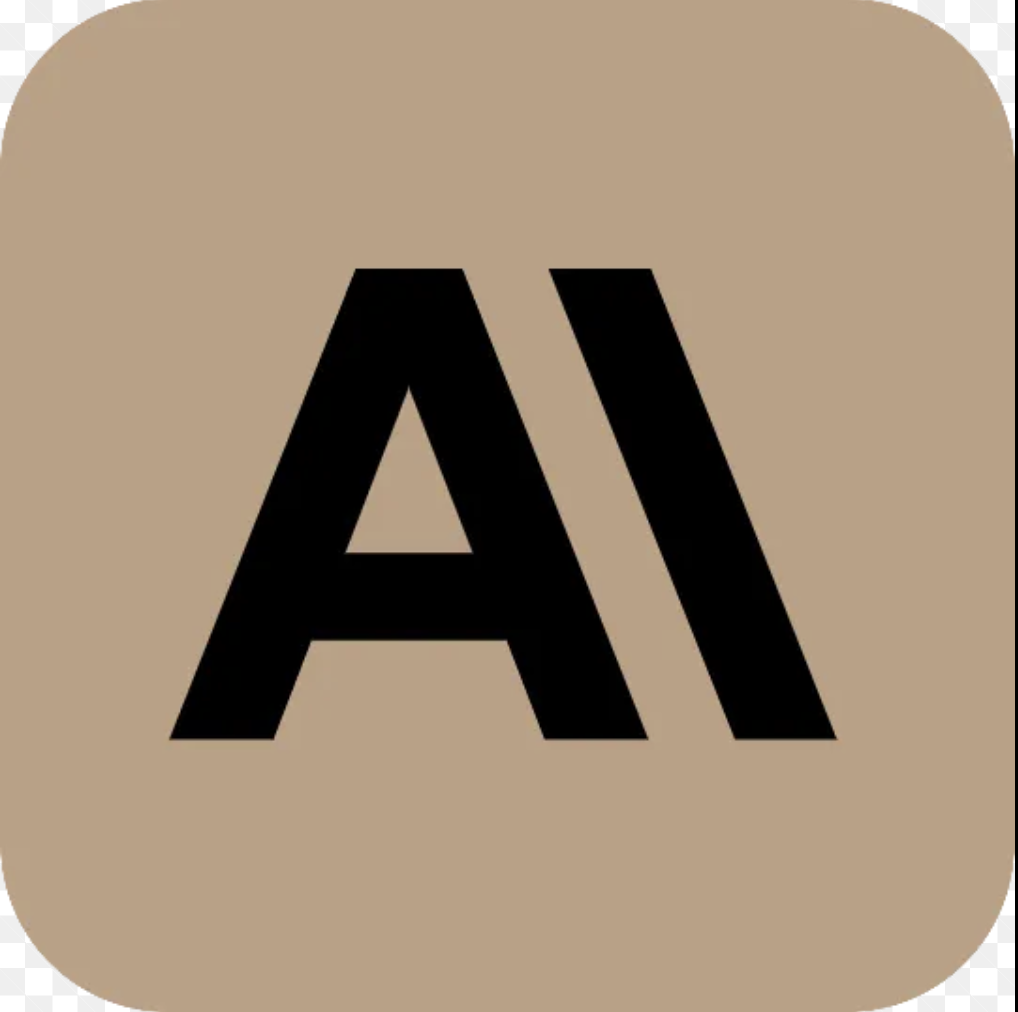}}~\textbf{Claude-Sonnet-3.7}%
}

\title{SciReplicate-Bench: Benchmarking LLMs in Agent-driven Algorithmic Reproduction from Research Papers}

\author{Yanzheng Xiang$^{1}$, Hanqi Yan$^{1}$, Shuyin Ouyang$^{1}$, Lin Gui$^{1}$, Yulan He$^{1,2}$  \\
$^1$King's College London, $^{2}$The Alan Turing Institute\\
\texttt{\{yanzheng.xiang,hanqi.yan,shuyin.ouyang,lin.1.gui,yulan.he\}@kcl.ac.uk} \\
}

\begin{document}

\ifcolmsubmission
\linenumbers
\fi

\maketitle
\begin{abstract}
This study evaluates large language models (LLMs) in generating code from algorithm descriptions in recent NLP 
papers. 
The task requires two key competencies: 
(1) algorithm comprehension: synthesizing information from papers and academic literature to understand implementation logic, and (2) coding expertise: identifying dependencies and correctly implementing necessary APIs.
To facilitate rigorous evaluation, we introduce \textbf{SciReplicate-Bench}, a benchmark of 100 tasks from 36 NLP papers published in 2024, featuring detailed annotations and comprehensive test cases.
Building on SciReplicate-Bench, we propose \textbf{Sci-Reproducer}, a dual-agent framework consisting of a Paper Agent that interprets algorithmic concepts from literature and a Code Agent that retrieves dependencies from repositories and implements solutions.
To assess algorithm understanding, we introduce \textit{reasoning graph accuracy}, which quantifies similarity between generated and reference reasoning graphs derived from code comments and structure.
For evaluating implementation quality, we employ \textit{execution accuracy}, \textit{CodeBLEU}, and repository dependency/API \textit{recall} metrics.
In our experiments, we evaluate various powerful non-reasoning and reasoning LLMs as foundational models.
The best-performing LLM using \ModelName~achieves only 39\% \textit{execution accuracy}, highlighting the benchmark's difficulty.
Our analysis identifies missing or inconsistent algorithm descriptions as key barriers to successful reproduction. 
 We make available our benchmark and code at \href{https://github.com/xyzCS/SciReplicate-Bench}{\faGithub\ GitHub} and project homepage at \href{https://xyzcs.github.io/scireplicate.github.io/}{\faLink\ Homepage}.
\end{abstract}
\section{Introduction}
The evolution of Large Language Models (LLMs) has ushered in a transformative era in scientific discovery, positioning them as powerful tools for streamlining research \citep{Gridach2025AgenticAF,buehler2024graphreasoning,Lu2024TheAS}, from idea generation to verification and publication writing. For instance, \citet{si2025can,gu2025interestingscientificideageneration} demonstrated how LLMs can be prompted to generate novel research ideas, while~\citet{10.1613/jair.1.12862,du-etal-2024-llms} explored their use in producing literature reviews for idea evaluation. Additionally, LLMs are increasingly integrated into tools like Semantic Scholar\footnote{\url{https://www.semanticscholar.org}}, Research Rabbit\footnote{\url{https://www.researchrabbit.ai/}}, and Undermind Research Assistant\footnote{\url{https://www.undermind.ai/}}, enhancing literature discovery, citation analysis, and knowledge synthesis. These advancements, both in research methodologies and practical applications, suggest that LLMs have the potential to assist across multiple stages of scientific discovery.

Among the aforementioned advancements in research acceleration, the ability of LLMs to correctly generate code for validating real-world scientific ideas is particularly noteworthy. Computational validation is crucial across many fields, yet researchers often face barriers due to limited coding expertise or inaccessible implementations. By converting scientific algorithm descriptions into executable code, LLMs could enhance reproducibility and accelerate scientific discovery. However, despite progress in LLM-based code generation, a significant gap remains in generating code directly from scholarly papers. \textbf{\textit{First}}, \textit{algorithm comprehension from scientific papers is challenging.} Research papers are characterized by their brevity, methodological rigor, and extensive citations, 
with critical details about algorithms often dispersed across multiple sections of the paper. Understanding these algorithms requires synthesizing information from internal references and external scholarly works. \textbf{\textit{Second}}, \textit{code repositories typically consist of multiple interdependent files and directories}. To implement an algorithm, LLMs must comprehensively examine file dependencies, identify reusable components, and correctly handle both internal dependencies and external APIs.

Despite the importance of automated scientific idea verification, there exists no dataset specifically designed to evaluate the ability of LLMs to reproduce real-world  algorithms proposed in peer-reviewed publications.
As shown in Table~\ref{tab:bench_compare}, there are several machine learning software engineering benchmarks primarily focused on evaluating algorithmic design or straightforward implementations, which are significantly less complex than the methods typically described in academic research papers.
For example, MLE-BENCH~\citep{Liu2023MLBenchEL} and MLAgentBench~\citep{Huang2023MLAgentBenchEL} utilize Kaggle competitions, where LLMs must develop and implement solutions based on provided task specifications. ML-BENCH~\citep{Chan2024MLEbenchEM} uses Machine Learning (ML) GitHub repositories to assess LLMs' text-to-code capabilities and test autonomous agents in task execution. 
\renewcommand{\arraystretch}{0.7}
\begin{table}[h]
    \centering
    \resizebox{1\textwidth}{!}{%
    \begin{tabular}{r|ccccl}
\toprule
    \textbf{Benchmark}  & \textbf{Paper Understanding}  & \textbf{Repo-Search} & \textbf{Test Case} & \textbf{Source} & \textbf{Task Types}\\
\midrule
MLE-BENCH  &  \XSolidBrush    & \XSolidBrush & \XSolidBrush & Kaggle & Algorithm design and code gen.\\
MLAgentBench & \XSolidBrush  & \Checkmark & \Checkmark &Kaggle & Algorithm design and code gen.\\
ML-BENCH & \XSolidBrush & \Checkmark & \Checkmark & Github & Code gen.\\
\midrule
\multirow{2}{*}{\DatasetName} & \multirow{2}{*}{\Checkmark} & \multirow{2}{*}{\Checkmark} & \multirow{2}{*}{\Checkmark} &  \multirow{2}{*}{Publications} & Replicate code for algorithms \\
& & & & & in real-world NLP publications. \\
\bottomrule
    \end{tabular}
    }
    \caption{Comparisons of different machine learning software engineering benchmarks.}
    \label{tab:bench_compare}
\end{table}

To address this gap, we developed \textbf{\DatasetName}, \textbf{the first benchmark specifically designed to evaluate LLMs' capabilities in code generation for reproducing research findings from academic papers}. It consists of 100 code reproduction tasks derived from 36 papers published in leading conferences in 2024. This recent publication window was deliberately chosen to minimize the risk of data leakage.
An overview of the task is illustrated in Figure~\ref{method}, with a concrete example provided in Figure~\ref{taskintro} in Appendix~\ref{FigureTable}.
The task consists of two main steps: \textbf{1. Algorithm understanding}.  LLMs must extract essential information from the paper, such as workflow details, algorithm descriptions, and hyperparameter values. \textbf{2. Code implementation}. 
LLMs then implement a function or method within a provided repository, using both the extracted information and the LaTeX representation of the algorithm from the paper.
We introduce \textbf{\ModelName}, a dual-agent system that combines a Paper Agent and a Code Agent to handle these two steps collaboratively and implement code for the target algorithm.

To rigorously assess LLM performance on this benchmark, we evaluate two dimensions corresponding to the aforementioned two steps: \emph{algorithm comprehension correctness} and \emph{code correctness}.
To evaluate algorithm comprehension, we introduce a \textbf{reasoning graph} to represent the reasoning logic behind algorithm reproduction.
Each node in the graph represents a code comment, which reflects a single reasoning step and is aligned with a specific segment of code.
Edges between nodes are defined based on data flow relationships across different code segments.
We compute the similarity between the generated reasoning graph and a reference graph to derive the \textit{reasoning graph accuracy}.
To evaluate code correctness, we employ established metrics including \textit{execution accuracy}~\citep{Rajkumar2022EvaluatingTT,Xiang2023GRAG}, \textit{CodeBLEU}~\citep{Ren2020CodeBLEUAM}, and \textit{recall} of intra/cross-file dependencies and APIs.

 Our work makes the following contributions:
\begin{itemize}[itemsep=0pt, topsep=-2pt,leftmargin=*]
    \item[] \textbf{Benchmarks}: \DatasetName, a benchmark of 100 algorithm reproduction tasks from recent NLP publications.
    \item[] \textbf{Metric}: We propose a novel \textit{reasoning graph accuracy} metric for evaluating algorithmic comprehension.
    \item[] \textbf{Approach}: \ModelName, a dual-agent framework combining paper understanding and code implementation.
    \item[] \textbf{Insights}: Comprehensive evaluation across state-of-the-art LLMs reveals four key findings: (i) the task remains highly challenging, with execution accuracy below 40\% for all models; (ii) reasoning models exhibit ``overthinking'' behavior~\citep{Cuadron2025TheDO,Sui2025StopOA}, over-relying on internal reasoning rather than utilizing available tools for information extraction; (iii) while LLMs demonstrate strong algorithmic comprehension, they struggle with practical implementation; and (iv) algorithm reproduction failures often stem from incomplete paper descriptions, which our \ModelName~effectively addresses.
\end{itemize}

\section{Related Work}
\label{relatedwork}
Our work lies at the intersection of AI for automating scientific discovery and LLM-based code generation.

\subsection{AI for Automating Scientific Discovery}
The application of LLMs to accelerate scientific research has emerged as a rapidly growing field with diverse approaches. Several studies have demonstrated the potential for comprehensive research automation through end-to-end AI systems. \citet{Schmidgall2025AgentLU,Lu2024TheAS} developed frameworks that integrate idea generation, experimental validation, and manuscript composition, with some AI-authored papers successfully passing workshop review processes~\citep{Yamada2025TheAS}. Complementary research has focused on the creative aspects of scientific inquiry, with \citet{Wang2023SciMONSI,Ghafarollahi2024SciAgentsAS,ONeill2025SparksOS} investigating LLMs' capacity for generating novel research hypotheses. 
Notably, recent evaluations by \citet{Gu2024LLMsCR, Kumar2024CanLL, Liu2025ResearchBenchBL,Si2024CanLG} suggest that AI-generated research concepts may occasionally exceed human-generated ideas in terms of novelty and originality.

Within computational disciplines where implementation validation is essential, LLMs have shown promise in algorithm design and code development tasks. 
Our proposed SciReplicate-Bench addresses a previously underexplored area: the automated reproduction of algorithms directly from academic publications. 
This represents a unique challenge at the intersection of scientific literature comprehension and executable code synthesis. 
While recent parallel efforts such as PaperBench~\citep{Starace2025PaperBenchEA} and Paper2CodeBench~\citep{Seo2025Paper2CodeAC} have tackled related problems by exploring full codebase reconstruction, these evaluation approaches rely substantially on manual assessment criteria and LLM-based correctness judgments, introducing potential inconsistencies and reliability concerns.
Our approach prioritizes objective evaluation through \textit{execution accuracy}, providing more rigorous validation than non-executable assessment methodologies~\citep{Wang2022ExecutionBasedEF, Chen2021EvaluatingLL}.

The broader ecosystem of computational reproducibility research includes specialized frameworks such as MLGym~\citep{Nathani2025MLGymAN} for baseline improvement, and evaluation benchmarks developed by \citet{Siegel2024COREBenchFT,Ren2023SuperBenchAS} that assess LLMs' ability to reproduce published experimental results using existing codebases.

\subsection{LLMs for Code Generation}
Code generation has emerged as a prominent application of LLMs, with benchmarks ranging from basic programming tasks~\citep{Chen2021EvaluatingLL,Jain2024LiveCodeBenchHA,Austin2021ProgramSW,Hendrycks2021MeasuringCC,Liu2022CodeGF} to realistic software engineering challenges like SWE-bench~\citep{Jimenez2023SWEbenchCL}, which uses actual repository pull requests. However, these benchmarks primarily target general software engineering rather than scientific algorithm reproduction.

Recent efforts have developed machine learning-specific benchmarks~\citep{Liu2023MLBenchEL,Huang2023MLAgentBenchEL,Chan2024MLEbenchEM}, but these typically involve implementing algorithms proposed by the models themselves or solving relatively straightforward tasks. They lack the depth of algorithmic understanding and rigorous paper analysis required for reproducing algorithms from peer-reviewed publications.

Despite advances in tool-augmented code generation~\citep{Schick2023ToolformerLM,Zhang2024CodeAgentEC,Zhang2024CodeAgentEC,Zhang2023ToolCoderTC}, no existing system specifically addresses the unique challenge of translating academic papers into executable code. Our Sci-Reproducer framework demonstrates the ability to comprehend academic publications and convert abstract algorithm descriptions into functional implementations.
\section{\DatasetName}

\paragraph{Overview}
\DatasetName~is designed to evaluate LLMs' ability to reproduce algorithms from academic papers, consisting of 100 tasks curated from 36 recent NLP publications with their corresponding open-source implementations.
The task categories are detailed in Figure~\ref{taskcate} in Appendix~\ref{Appen:taskcate}.
The benchmark focuses on repository-level code generation, where each task is centered around implementing a specific function or class method.
As illustrated in Figure~\ref{Appen:datasetintro} in Appendix~\ref{FigureTable}, each task comprises nine components, which can be categorized into three groups corresponding to code generation, evaluation, and analysis, respectively.

For code generation, the following components are provided as inputs to LLMs:
 \begin{enumerate}[itemsep=0pt, topsep=-2pt,leftmargin=*]
     \item[] \textit{Function signature}: 
    the definition of the target function, including detailed descriptions of its input and output variables.

    \item[] \textit{Algorithm Description}: 
    The LaTeX code description of the target algorithm, typically located within a subsection or paragraph of the target paper.
    
    \item[] \textit{Literature context}:  
    the original paper along with its cited references, providing broader conceptual context.
    
    \item[] \textit{Repository context}:  
    all source files and code in the repository that inform or support the target implementation.
\end{enumerate}

For evaluation, the following components are provided for code execution and metrics calculation:

 \begin{enumerate}[itemsep=0pt, topsep=-2pt,leftmargin=*]
    \item[] \textit{Reference implementation}:
    ground-truth code serving as the reference for \textit{CodeBLEU} evaluation.
    
    \item[] \textit{Reasoning graph annotations}:
    structured representations of the algorithmic logic and implementation flow, enabling assessment of \textit{reasoning graph accuracy}.
    
    \item[] \textit{Dependency annotations}:
    comprehensive documentation of internal dependencies, cross-file relationships, and external API usage for computing \textit{recall} metrics across all dependency categories.
    
    \item[] \textit{Test environment}: 
    isolated Python execution environment containing validation cases and automated verification scripts for assessing implementation correctness.
\end{enumerate}
 
To enable further analysis of the underlying causes of LLM failures, the benchmark includes:
 \begin{enumerate}[itemsep=0pt, topsep=-2pt,leftmargin=*]
     \item[] \textit{Missing/Mismatch Information}: the LaTeX description of the algorithm may omit certain implementation details, which could either appear elsewhere in the paper or be entirely absent. 
     We also annotate  mismatches between the paper description and the reference implementation.
 \end{enumerate}
     
\paragraph{Task Definition}
Based on \DatasetName, an LLM is given the algorithm description, function signature, literature context, and repository context as input. 
The LLM is asked to output a function that implements the target algorithm.
\paragraph{Benchmark Construction} The benchmark construction process comprises four key steps: paper selection, Python environment setup, documentation, and verification suite preparation.
To mitigate the risk of data leakage, we selected papers published in 2024 that provide publicly available code repositories. During the annotation process, each repository was refactored to isolate the core algorithm into a standalone function, and all sources of randomness were removed to ensure reproducibility and prevent leakage.
On average, annotating each paper requires approximately 12 hours.
Details of the annotation process are provided in Appendix~\ref{AnnotationProcess}.

\subsection{Evaluation Metrics}
\subsubsection{Evaluating Algorithm Comprehension}
We propose the \textit{reasoning graph accuracy} metric to evaluate how well LLMs understand the logic and implementation of algorithms.
During code generation, LLMs are prompted to insert specially formatted, non-overlapping, non-nested comments that mark reasoning steps derived from the algorithm’s LaTeX code (The prompt can be found in Figure~\ref{PromptCodeGen}).
We then construct a reasoning graph $G = \{N, E\}$ (illustrated in Figure~\ref{Appen:datasetintro}), modeled as a Directed Acyclic Graph (DAG). Each node $n_i = \langle w_i, c_i \rangle, n_i \in N$ represents a reasoning step with a comment $w_i$ and corresponding code snippet $c_i$. An edge $e_i = \langle n_i, n_j \rangle, e_i \in E$ is added if a variable used in $c_j$ is defined or last modified in $c_i$.
To compute the \textit{reasoning graph accuracy}, we compare the generated graph $G_g$ with the reference graph $G_r$ via node and edge matching:
\begin{enumerate}[itemsep=0pt, topsep=-2pt,leftmargin=*]
    \item[] \textbf{Node matching:} comments from $G_r$ and $G_g$ are passed to GPT-4o, which maps each reference node to one or more nodes in the generated graph. A node in $G_r$ is considered matched if it has at least one corresponding node in $G_g$. 
    The prompt template used for this process is available in Figure~\ref{PromptAlign}.
    \item[] \textbf{Edge matching:} for each reference edge $e_r = \langle n_r^i, n_r^j \rangle$, if both endpoint nodes have corresponding matches in $G_g$, we apply Breadth-First Search(BFS) to verify whether a corresponding edge exists in $G_g$.
\end{enumerate}
The \textit{reasoning graph accuracy} $S_r$ is computed as:
\begin{equation}
\small
    S_r = \sum_{n_i}^{n_i\in N_m}s^n_i + \sum_{e_j}^{e_j\in E_m}s^e_j.
\end{equation}
where $N_m$ and $E_m$ denote the sets of matched nodes and edges, respectively, and $s_i^n$ and $s_j^e$ represent their corresponding significance scores.
Node significance is determined by the complexity of its corresponding code segment, measured by the number of variable definitions and usages, function calls, arithmetic operations, and lines of code, then normalized across the reference graph.
Edge significance is calculated as the product of the significance scores of its connected nodes, followed by normalization.

\subsubsection{Evaluating Code Generation}
For assessing coding ability, we use the following evaluation metrics:
\begin{itemize}[itemsep=0pt, topsep=-2pt,leftmargin=*]
    \item \textit{Execution accuracy}~\citep{Xiang2023GRAG,Zhang2024HFDHF,Long2022NL2SQLGW}: we integrate the generated code into the repository and execute it to obtain results. If all test cases match the reference results, we consider the code correct. 
    \item \textit{CodeBLEU}~\citep{Ren2020CodeBLEUAM}: this metric evaluates how similar the generated code is to reference code by using the traditional BLEU metric~\citep{Papineni2002BleuAM} while incorporating syntactic information through abstract syntax trees (AST) and semantic understanding via data-flow graphs (DFG).
    \item \textit{Recall}~\citep{Li2024DevEvalAM}: we calculate recall scores specifically for intra-file dependencies, cross-file dependencies, and external APIs.
\end{itemize}

\section{\ModelName}

\begin{figure}[t]
 	\centering
 	\includegraphics[width=0.65\textwidth]{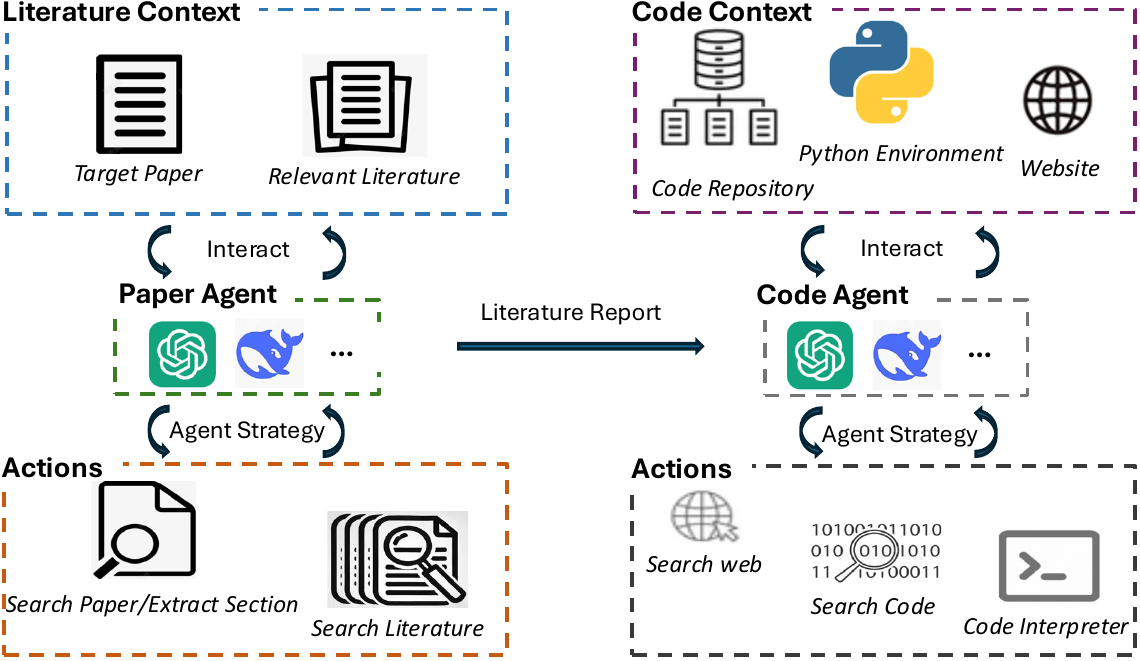} 
 	\caption{\label{method}
 	Overview of the task and the proposed \ModelName~framework. The task involves algorithm understanding and code implementation, handled by a Paper Agent and a Code Agent operating in separate contexts with specialized actions.}
\end{figure}

\renewcommand{\arraystretch}{0.7}
\begin{table*}[h]
\centering
\resizebox{0.85\linewidth}{!}{
\begin{tabular}{l|l|l}
\toprule
\textbf{Action Name} & \textbf{Input} & \multicolumn{1}{c}{\textbf{Observation}}\\
\midrule
\multicolumn{3}{c}{\cellcolor{gray!30} \textbf{Paper Agent}} \\ 
\midrule
\multirow{2}{*}{SearchPaper} & \multirow{2}{*}{Query} & The retrieved response from the target paper in relation to \\
& &the query.\\
\midrule
SearchSection & Section ID & The entire content of a section based on the section label.\\
\midrule
\multirow{2}{*}{SearchLiterature} & \multirow{2}{*}{Paper ID, query} & The answer to the query searched from the literature (identified \\
& &   by Paper ID).\\
\midrule
\multicolumn{3}{c}{\cellcolor{gray!30} \textbf{Code Agent}} \\ 
\midrule
SearchCode & Name & The definition of a specific code element in repository.\\
\midrule
SearchFile & Name & The content of a certain file in repository.\\
\midrule
SearchWeb & Query & The information obtained from the website.\\
\midrule
Compiler & code & The feedback from the compiler after executing the code. \\
\bottomrule
\end{tabular}}
\caption{\label{action}
	The pre-defined actions for the Paper Agent and the Code Agent.}
\end{table*}
 
To address this task, we introduce \ModelName~\footnote{A video demonstration showcasing Sci-Reproducer's capabilities is provided at \url{https://youtu.be/qcSIMgyehjE}}, a dual-agent framework designed for scientific paper algorithm replication.
As illustrated in Figure~\ref{method}, \ModelName~ comprises a Paper Agent and a Code Agent that collaboratively work to replicate algorithms described in a given paper.
The predefined actions employed by the agents are summarized in Table~\ref{action}, with implementation details provided in Appendix~\ref{Appen:tool}.

\subsection{Paper Agent}
Based on the provided algorithm description, the Paper Agent systematically retrieves contextual information from the literature context to support algorithmic understanding.
Due to the input length limitations of LLMs, it is infeasible to input entire paper along with their associated literature. 
Consequently, the Paper Agent must selectively extract pertinent information, following a strategy akin to Retrieval Augmented Generation (RAG)~\citep{Wang2024GARLICLD,Sarthi2024RAPTORRA}.
The Paper Agent incrementally builds an understanding of the target algorithm by executing predefined actions to query the literature context. 
To facilitate this process, we adopt ReAct~\citep{Yao2022ReActSR} as the agent strategy, which enables seamless integration of action execution with intermediate reasoning steps.

After the Paper Agent concludes that all necessary information has been collected, it generates a comprehensive literature report comprising key findings that fill in the missing components of the target algorithm's LaTeX source.
An example of the literature report is shown in Figure~\ref{PaperAgentOutput}.
This report subsequently serves as a crucial input for the Code Agent.
The prompt used to guide the Paper Agent is provided in Figure~\ref{PromptPaper}.

\subsection{Code Agent}
Informed by the algorithm description, literature report, and code context, the Code Agent searches the repository to locate essential dependencies required for implementation.
It can also browse websites for additional information and use a compiler to test and iteratively debug the code, ensuring proper execution by identifying and fixing syntax errors.
The prompt for the Code Agent is provided in Figure~\ref{PromptCode}.

\section{Experiments}

\renewcommand{\arraystretch}{0.7}
\begin{table*}[t]
\centering
\resizebox{0.85\linewidth}{!}{
\begin{tabular}{l|c|c|c|c|c|c}
    \toprule
    \multirow{2}{*}{Approach} & \multirow{2}{*}{Exe Acc($\uparrow$)} & \multirow{2}{*}{CodeBLEU($\uparrow$)} & \multirow{2}{*}{RG Acc($\uparrow$)} & \multicolumn{3}{c}{Recall} \\
    &  & &  & Intra-File($\uparrow$) & Cross-File($\uparrow$) & API($\uparrow$) \\
    \midrule
    \multicolumn{7}{c}{ \gptmini} \\
    \midrule
    No Agent & 0.030 & 0.238 & 0.694 & 0.012 & 0.000  & 0.217 \\
    Paper Agent  & 0.040 & 0.246 & 0.717 & 0.024 & 0.000 & 0.251 \\
    Code Agent  & 0.140 & 0.279 & 0.738 & 0.565 & 0.364 & 0.328 \\
    \ModelName & \textbf{0.170} & \textbf{0.303} & \textbf{0.741} & \textbf{0.576} & \textbf{0.364} & \textbf{0.362} \\
    \midrule
    \multicolumn{7}{c}{\gpttwo} \\
    \midrule
    No Agent & 0.040 & 0.259 & 0.705 & 0.059 & 0.000 & 0.281 \\
    Paper Agent  & 0.020 & 0.263 & 0.724 & 0.023 & 0.000 & 0.298\\
    Code Agent  & 0.260 & 0.325 & 0.748 & 0.682 & 0.576 & \textbf{0.421} \\
    \ModelName & \textbf{0.270} & \textbf{0.329} & \textbf{0.751} & \textbf{0.688} & \textbf{0.636} & 0.417 \\
    \midrule
    \multicolumn{7}{c}{\claude} \\
    \midrule
    No Agent & 0.070 & 0.282 & 0.725 & 0.094 & 0.091 & 0.362 \\
    Paper Agent  & 0.050 & 0.291 & 0.727 & 0.082 & 0.091 & 0.379\\
    Code Agent  & 0.320 & 0.394 & 0.764 & 0.765 & 0.545 & 0.545 \\
    \ModelName & \textbf{0.390} & \textbf{0.401} & \textbf{0.773} & \textbf{0.776} & \textbf{0.636} & \textbf{0.626} \\
    \midrule
    \multicolumn{7}{c}{\gemini} \\
    \midrule
    No Agent & 0.070 & 0.275 & 0.688 & 0.071 & 0.000 & 0.294 \\
    Paper Agent  & 0.040 & 0.278 & 0.699 & 0.082 & 0.000 & 0.332\\
    Code Agent  & 0.220 & 0.323 & 0.708 & 0.553 & 0.212 & 0.426 \\
    \ModelName & \textbf{0.250} & \textbf{0.346} & \textbf{0.727} & \textbf{0.588} & \textbf{0.333} & \textbf{0.455} \\
    \midrule
    \multicolumn{7}{c}{\deepseek} \\
    \midrule
    No Agent & 0.030 & 0.260 & 0.712 & 0.012 & 0.061 & 0.272 \\
    Paper Agent  & 0.050 & 0.275 & 0.732 & 0.012 & 0.030 & 0.306\\
    Code Agent & 0.210 & 0.312 & 0.736 & 0.482 & 0.182 & 0.383 \\
    \ModelName & \textbf{0.220} & \textbf{0.334} & \textbf{0.738} & \textbf{0.565} & \textbf{0.333} & \textbf{0.443} \\
    \midrule
    \multicolumn{7}{c}{\olow~\thinking} \\
    \midrule
    No Agent & 0.080 & 0.259 & 0.758 & 0.035 & 0.000 & 0.323 \\
    Paper Agent  & 0.050 & 0.262 & 0.729 & 0.035 & 0.000 & 0.315 \\
    Code Agent  & 0.150 & 0.278 & \textbf{0.803} & 0.306 & 0.000 & \textbf{0.348} \\
    \ModelName & \textbf{0.180} & \textbf{0.280} & 0.771 & \textbf{0.376} & \textbf{0.121} & 0.328 \\
    \midrule
    \multicolumn{7}{c}{\omedium~\thinking} \\
    \midrule
    No Agent & 0.040 & 0.263 & 0.729 & 0.035 & 0.000 & 0.336 \\
    Paper Agent  & 0.060 & 0.263 & 0.726 & 0.047 & 0.000 & 0.319\\
    Code Agent  & 0.220 & \textbf{0.289} & 0.749 & \textbf{0.376} & 0.030 & \textbf{0.404} \\
    \ModelName & \textbf{0.240} & 0.283 & \textbf{0.758} & 0.341 & \textbf{0.061} & 0.362 \\
    \midrule
    \multicolumn{7}{c}{\ohigh~\thinking} \\
    \midrule
    No Agent & 0.070 & 0.269 & 0.718 & 0.047 & 0.000 & 0.345 \\
    Paper Agent  & 0.070 & 0.267 & 0.751 & 0.035 & 0.000 & 0.366\\
    Code Agent  & 0.160 & 0.277 & \textbf{0.773} & 0.165 & \textbf{0.152} & \textbf{0.374} \\
    \ModelName & \textbf{0.160} & \textbf{0.283} & 0.763 & \textbf{0.294} & 0.091 & 0.357 \\
    \bottomrule
\end{tabular}}
\caption{\label{MainResults}
	Performance evaluation on the \DatasetName~benchmark. Models with \protect \thinking notation indicate \textbf{reasoning LLMs}. ``Exe Acc'' represents \textit{execution accuracy} while ``RG Acc'' indicates \textit{reasoning graph accuracy}. 
    }
\end{table*}
We evaluate \ModelName~on the \DatasetName~ benchmark using 7 advanced LLMs, including five non-reasoning LLMs: GPT-4o-mini~\citep{GPT-4o-mini}, GPT-4o~\citep{GPT-4o}, Claude-Sonnet-3.7~\citep{Claude-Sonnet-3.7}, Gemini-2.0-Flash~\citep{Gemini-2.0-Flash}, and Deepseek-V3~\citep{DeepSeekAI2024DeepSeekV3TR}, and different versions of the reasoning models O3-mini~\citep{o3-mini}, i.e., three different levels of reasoning intensity. 
For the \textit{reasoning graph accuracy} metric, node matching is performed using GPT-4o, which may introduce some randomness. To reduce this variability, we set the temperature to 0 and top-p to 1, ensuring more deterministic generation. The calculation is repeated three times, and we report the average score as the final result.

\subsection{Results on \DatasetName}
Table~\ref{MainResults} displays \ModelName's evaluation results and contributions of Code/Paper Agent. The ``No Agent''directly prompts the LLM to generate code based solely on the algorithm description and function signature. 
``No Paper Agent'' allows the LLM to use Code Agent actions, but restricts access to Paper Agent actions.
``No Code Agent'' grants access to Paper Agent actions but blocks Code Agent capabilities. 
The results offer key insights, discussed in the following.

\textbf{LLMs struggles on \DatasetName}
Most LLMs perform poorly, achieving less than 0.1 \textit{execution accuracy} without using the agent to examine literature and repository contexts.
With enhancement of \ModelName, these LLMs show notable improvements, with an average increase of 0.181 in execution ACC and 0.057 in \textit{CodeBLEU}, although even the best-performing model, Claude-Sonnet-3.7, only achieved 0.390 \textit{execution accuracy}. This highlights the exceptional challenge presented by our \DatasetName.

\textbf{LLMs can comprehend algorithm logic}
LLMs demonstrate strong algorithmic comprehension capabilities, as evidenced by \textit{reasoning graph accuracy scores} averaging 0.716 even without agent assistance. The addition of individual agents provides modest but consistent improvements: the Paper Agent increases understanding by 0.009 on average, while the Code Agent contributes a larger gain of 0.036. Combined agent deployment yields a cumulative improvement of 0.037.
These enhancements stem from complementary mechanisms: the Paper Agent strengthens theoretical comprehension by gathering relevant contextual information from academic literature, while the Code Agent facilitates practical understanding through extraction of pertinent code patterns and dependency structures from repositories.

\textbf{LLMs face challenges with actual implementation}
Although LLMs are capable of understanding algorithms, their performance in code generation remains suboptimal.
Despite using \ModelName, the average \textit{execution accuracy} remains low at 0.235, with a \textit{CodeBLEU} score of 0.320.

\textbf{Accurate dependency and API identification is crucial for code implementation}
Effectively recognizing and leveraging dependencies from the source repository and external APIs is essential for accurate code implementation. 
The integration of Code Agent led to substantial gains in \textit{recall}
with average increases of 0.441, 0.239, and 0.100, respectively, compared to cases without the agent.
With \ModelName, Claude-Sonnet-3.7 attains the highest \textit{execution accuracy} of 0.390, with the highest \textit{recall} for intra/cross file dependency and API usage, at 0.776, 0.636, and 0.626 respectively.

\textbf{Overthinking leads to limited improvement in reasoning LLMs}
Reasoning LLMs exhibit more modest performance gains when using \ModelName. While they achieve an average \textit{execution accuracy} improvement of 0.13, non-reasoning models demonstrate substantially larger gains of 0.212. This pattern extends to \textit{recall} metrics, where reasoning LLMs show improvements of 0.243, 0.061, and 0.041 respectively, compared to non-reasoning LLMs' more pronounced gains of 0.560, 0.345, and 0.135. We attribute this performance gap to the ``overthinking" phenomenon~\citep{Cuadron2025TheDO,Sui2025StopOA}, where excessive internal reasoning impedes effective action execution, a limitation that we examine in detail in the following subsection.

\subsection{Tool Usage Analysis}
\begin{figure}[t]
 	\centering
 \includegraphics[width=1\textwidth, height=0.4\textwidth]{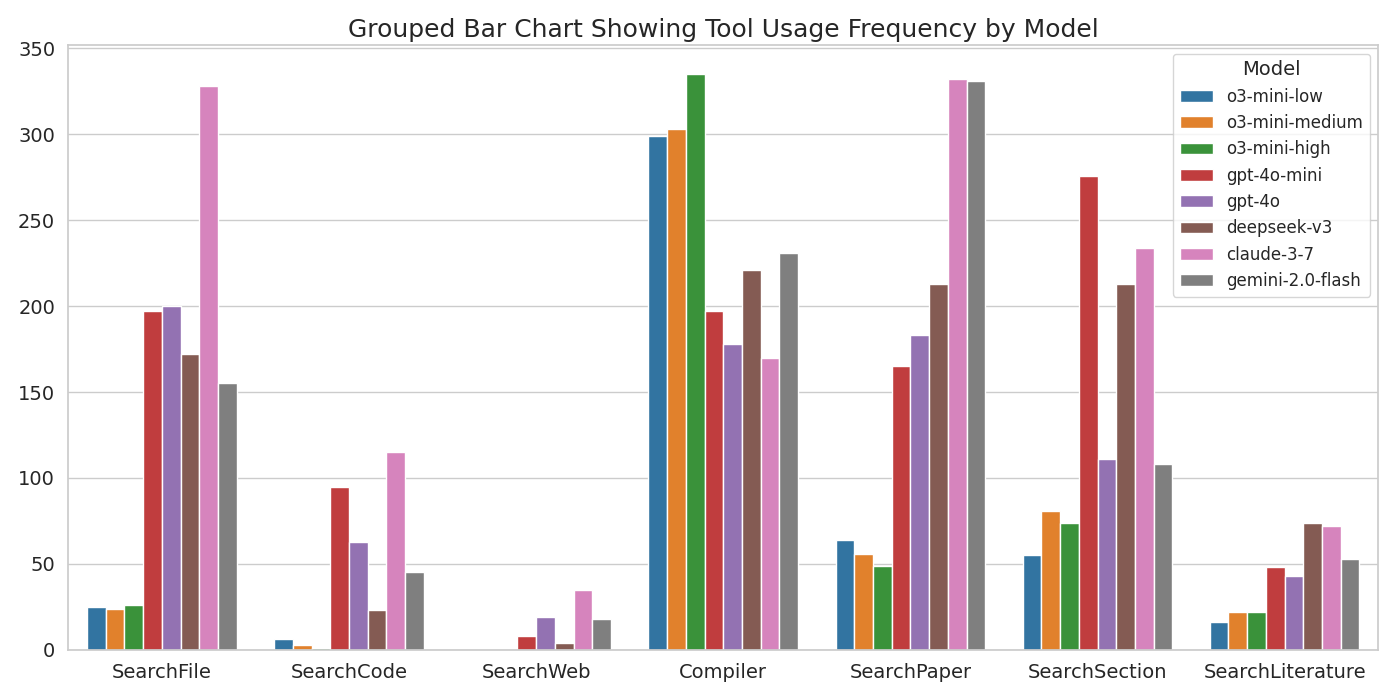} 
 	\caption{\label{tooluse}
 	A grouped bar chart illustrating the frequency of tool usage by different models. The x-axis represents various actions, while the y-axis indicates the total number of times each tool was used on this dataset.}
 \end{figure}

Figure~\ref{tooluse} presents the number of times each LLM invokes actions on the full dataset with \ModelName. 
We observe the following: 
\begin{itemize}[itemsep=0pt, topsep=-2pt,leftmargin=*]
    \item For code-related actions, reasoning LLMs demonstrate limited tool usage, employing ``SearchFile'', ``SearchCodeItem'', and ``SearchWeb'' only 25.0, 3.3, and 0.0 times on average, respectively. Non-reasoning LLMs use these same actions far more extensively, with averages of 210.4, 68.2, and 16.8 times respectively. This disparity reveals a fundamental behavioral difference: reasoning models favor internal deliberation over external information gathering. Conversely, reasoning LLMs invoke ``Compiler'' more frequently, indicating they require more debugging iterations due to inadequate contextual information gathering. This over-reliance on internal reasoning undermines performance: advanced models like o3-mini-high and o3-mini-low achieve \textit{execution accuracy} comparable to GPT-4o-mini, negating their theoretical computational advantages.
    
    \item Paper-related actions exhibit a similar pattern. Reasoning LLMs use ``SearchPaper", ``SearchSection", and ``SearchLiterature" an average of 56.3, 70.0, and 20.0 times respectively, while non-reasoning LLMs demonstrate substantially higher usage at 244.8, 188.4, and 58.0 times respectively. 
    Additionally, we observe a clear preference for target paper extraction over external literature consultation. Actions targeting the primary paper (``SearchPaper'' and ``SearchSection'') are invoked 174.1 and 144 times on average, significantly more than ``SearchLiterature'' which accesses related works only 43.8 times.
\end{itemize}

\renewcommand{\arraystretch}{0.7}

\subsection{Error Analysis}

\subsubsection{Syntax Errors}
Table~\ref{Syntax} shows the syntax error rates for each model across different configurations. Without the Code Agent, syntax errors occurred at rates of 80.3\% (``NoAgent'') and 83.3\% (``Paper Agent''). 
After implementing the Code Agent, these error rates dropped significantly to 29.4\% (``Code Agent'') and 24.9\% (``\ModelName''). 
The remaining syntax errors mainly result from incorrectly using repository dependencies. This occurs because our approach, unlike human developers, cannot dynamically access runtime information through a compiler during the code generation process.

\subsubsection{Logic Errors}
Another issue stems from differences in implementation logic, which can be broadly categorized into: (1) discrepancy in algorithm implementation that result in differing outputs, and (2) missing or mismatch information in the algorithm descriptions in the paper compared to the actual code.

\textbf{Implementation discrepancy}
An algorithm may have multiple valid implementation approaches. For example, the cross-entropy loss function can be implemented by directly invoking the PyTorch API ``torch.nn.CrossEntropy'' or by manually coding it from scratch. 
Such implementation choices may introduce subtle differences that lead to variations in the final output of the function.\\
\textbf{Missing/Mismatched information in algorithm description}
\renewcommand{\arraystretch}{0.7}
\begin{table*}[t]
\centering
\resizebox{0.9\linewidth}{!}{
\begin{tabular}{l|c|c|c|c|c|c}
    \toprule
    \multirow{2}{*}{\textbf{Model}~(\ModelName)} & \multirow{2}{*}{\textbf{Exe Acc}($\uparrow$)} & \multirow{2}{*}{\textbf{CodeBLEU}($\uparrow$)} & \multirow{2}{*}{\textbf{RG Acc}($\uparrow$)} & \multicolumn{3}{c}{\textbf{Recall}} \\
    &  & &  & Intra-File($\uparrow$) & Cross-File($\uparrow$) & API($\uparrow$) \\
    \midrule
    \gptmini & 0.220 & 0.316 & 0.809 & 0.588 & 0.485 & 0.409 \\
    \deepseek & 0.470 & 0.378 & 0.834 & 0.682 & 0.424 & 0.609\\
    \olow~\thinking & 0.220 & 0.292 & 0.850 & 0.259 & 0.091 & 0.460\\
    \bottomrule
\end{tabular}}
\caption{\label{Missing}
    Experimental Results when missing/mismatched information is regard as external input in the prompt.
}
\end{table*}
Algorithmic descriptions in research papers often lack concrete implementation details, and in certain cases, the provided code may exhibit minor discrepancies compared to the descriptions in the paper.
We manually compared the implementation code of all tasks in the dataset with their descriptions in the papers to identify missing or mismatch information. 
We then provided this information as additional input and apply \ModelName~framework on three LLMs.
The Results is shown in Table~\ref{Missing},
regarding to Execution Acc, the performance for GPT-4o-mini, Deepseek-V3 and O3-mini-low improved 0.050, 0.250 and 0.040 respectively.
The missing information can be divided into four categories:
\begin{itemize}[itemsep=0pt, topsep=-2pt,leftmargin=*]
    \item Hyperparameters and configurations: descriptions of target algorithms in papers often omit specific hyperparameter settings, such as the batch size.
    \item Numerical stability techniques: standard techniques for ensuring numerical stability, such as handling division by zero.
    \item Implementation logic: common implementation practices and model design choices, such as data splitting protocols.
    \item Coding strategy: practical programming techniques that enhance implementation efficiency and reliability, such as early stopping criteria.
\end{itemize}
More examples for each category can be found in Table~\ref{tab:miss} in Appendix~\ref{FigureTable}.
As for mismatched information, it occurs far less frequently compared to missing information, and its categories largely overlap with those mentioned above.

To mitigate the widespread issues of missing and mismatched information, the first category can generally be addressed by referencing the original research paper and related literature, or by inspecting the code repository for explicit configurations. 
However, addressing the other three categories requires familiarity with general machine learning coding conventions, thus necessitating that the LLMs identify and utilize implementation patterns from comparable algorithms to enhance code quality. 
Future research may improve performance by incorporating implementation insights from similar algorithms through techniques such as in-context learning~\citep{Zhou2023TheMO,Xiang2024AddressingOS}, and by leveraging real-time compiler feedback to infer precise variable values.

\section{Conclusion}
We evaluate LLMs’ ability to replicate algorithms described in recent NLP papers.
To support this, we introduce \DatasetName, a benchmark with rich annotations, and \ModelName, a dual-agent framework for bridging algorithm understanding and code generation.
We assess performance using \textit{reasoning graph accuracy} and standard implementation metrics.
Results show the task is highly challenging, with failures largely caused by missing or inconsistent algorithm descriptions.

\section{Acknowledgements}
This work was supported in part by the UK Engineering and Physical Sciences Research Council (EPSRC) through a Turing AI Fellowship (grant no. EP/V020579/1, EP/V020579/2).
We thank the authors of the selected papers for making their code openly available.

\bibliography{colm2025_conference}
\bibliographystyle{colm2025_conference}

\appendix
\section*{Appendix} 
\setcounter{table}{0}
\renewcommand{\thetable}{A\arabic{table}}
\setcounter{figure}{0}
\renewcommand{\thefigure}{A\arabic{figure}}

\section{Details of the Annotation Process}

  \begin{table}[h]
\vspace{-1em} 
\centering
\resizebox{\textwidth}{!}{  
\begin{tabular}{p{14.5cm}|c}
    \toprule
    \textbf{Title} & \textbf{Conference} \\
    \midrule
    1. From Zero to Hero: Cold-Start Anomaly Detection~\citep{Reiss2024FromZT} & Findings of ACL 2024\\
    \midrule
    2. Addressing Order Sensitivity of In-Context Demonstration Examples in Causal Language Models~\citep{Xiang2024AddressingOS} & Findings of ACL 2024 \\
    \midrule
    3. Breaking the Ceiling of the LLM Community by Treating Token Generation as a Classification for Ensembling~\citep{Yu2024BreakingTC} & Findings of EMNLP 2024 \\
    \midrule
    4. Simple but Effective Compound Geometric Operations for Temporal Knowledge Graph Completion~\citep{Ying2024SimpleBE} & ACL 2024 \\
    \midrule
    5. When is Tree Search Useful for LLM Planning? It Depends on the Discriminator~\citep{Chen2024WhenIT} & ACL 2024 \\
    \midrule
    6. Functional Overlap Reranking for Neural Code Generation~\citep{To2023FunctionalOR} & Findings of ACL 2024\\
    \midrule
    7. Beyond Single-Event Extraction: Towards Efficient Document-Level Multi-Event Argument Extraction~\citep{Liu2024BeyondSE} & Findings of ACL 2024 \\
    \midrule
    8. Unifying Dual-Space Embedding for Entity Alignment via Contrastive Learning~\citep{Wang2024UnifyingDE} & COLING 2025 \\
    \midrule
    9. Exploring Concept Depth: How Large Language Models Acquire Knowledge and Concepts at Different Layers?~\citep{Jin2024ExploringCD} & COLING 2025 \\
    \midrule
    10. TRANSMI: A Framework to Create Strong Baselines from Multilingual Pretrained Language Models for Transliterated Data~\citep{Liu2024TransMIAF} & COLING 2025 \\
    \midrule
    11. Enhancing Knowledge Distillation of Large Language Models through Efficient Multi-Modal Distribution Alignment~\citep{Peng2024EnhancingKD} & COLING 2025 \\
    \midrule
    12. Document-level Claim Extraction and Decontextualisation for Fact-Checking~\citep{Deng2024DocumentlevelCE} & ACL 2024 \\
    \midrule
    13. IRCAN: Mitigating Knowledge Conflicts in LLM Generation via Identifying and Reweighting Context-Aware Neurons~\citep{Shi2024IRCANMK} & Neurips 2024 \\
    \midrule
    14. RouterDC: Query-Based Router by Dual Contrastive Learning for Assembling Large Language Models~\citep{Chen2024RouterDCQR} & Neurips 2024 \\
    \midrule
    15. Unsupervised Homography Estimation on Multimodal Image Pair via Alternating Optimization~\citep{Song2024UnsupervisedHE} & Neurips 2024 \\
    \midrule
    16. RAPTOR: Recursive Abstractive Processing for Tree-Organized Retrieval~\citep{Sarthi2024RAPTORRA} & ICLR 2024 \\
    \midrule
    17. Less is KEN: a Universal and Simple Non-Parametric Pruning Algorithm for Large Language Models~\citep{Mastromattei2024LessIK} & Findings of ACL 2024\\
    \midrule
    18. Adaptive Contrastive Search: Uncertainty-Guided Decoding for Open-Ended Text Generation~\citep{Arias2024AdaptiveCS} & Findings of EMNLP 2024 \\
    \midrule
    19. MiniCheck: Efficient Fact-Checking of LLMs on Grounding Documents~\citep{Tang2024MiniCheckEF} & EMNLP 2024 \\
    \midrule
    20. Nearest Neighbor Normalization Improves Multimodal Retrieval~\citep{Chowdhury2024NearestNN} & EMNLP 2024 \\
    \midrule
    21. Neuron-Level Knowledge Attribution in Large Language Models~\citep{Yu2023NeuronLevelKA} & EMNLP 2024 \\
    \midrule
    22. RaTEScore: A Metric for Radiology Report Generation~\citep{Zhao2024RaTEScoreAM} & EMNLP 2024 \\
    \midrule
    23. GREEN: Generative Radiology Report Evaluation and Error Notation~\citep{Ostmeier2024GREENGR} & Findings of EMNLP 2024 \\
    \midrule
    24. Style-Specific Neurons for Steering LLMs in Text Style Transfer~\citep{Lai2024StyleSpecificNF} & EMNLP 2024 \\
    \midrule
    25. Language-Specific Neurons: The Key to Multilingual Capabilities in Large Language Models~\citep{Tang2024LanguageSpecificNT} & ACL 2024 \\
    \midrule
    26. Reasoning Paths Optimization: Learning to Reason and Explore From Diverse Paths~\citep{Chia2024ReasoningPO} & Findings of EMNLP 2024 \\
    \midrule
    27. Lifelong Knowledge Editing for LLMs with Retrieval-Augmented Continuous Prompt Learning~\citep{Chen2024LifelongKE} & EMNLP 2024 \\
    \midrule
    28. NeuroMax: Enhancing Neural Topic Modeling via Maximizing Mutual Information and Group Topic Regularization~\citep{Pham2024NeuroMaxEN} & Findings of EMNLP 2024 \\
    \midrule
    29. Bridging Local Details and Global Context in Text-Attributed Graphs~\citep{Wang2024BridgingLD} & EMNLP 2024 \\
    \midrule
    30. Advancing Adversarial Suffix Transfer Learning on Aligned Large Language Models~\citep{Liu2024AdvancingAS} & EMNLP 2024 \\
    \midrule
    31. SafeDecoding: Defending against Jailbreak Attacks via Safety-Aware Decoding~\citep{Xu2024SafeDecodingDA} & ACL 2024 \\
    \midrule
    32. MaskLID: Code-Switching Language Identification through Iterative Masking~\citep{Kargaran2024MaskLIDCL} & ACL 2024 \\
    \midrule
    33. Towards Robust and Generalized Parameter-Efficient Fine-Tuning for Noisy Label Learning~\citep{Kim2024TowardsRA} & ACL 2024 \\
    \midrule
    34. Learning to Maximize Mutual Information for Chain-of-Thought Distillation~\citep{Chen2024LearningTM} & Findings of ACL 2024 \\
    \midrule
    35. GLiNER: Generalist Model for Named Entity Recognition using Bidirectional Transformer~\citep{Zaratiana2023GLiNERGM} & NAACL 2024\\
    \midrule
    36. End-to-End Beam Retrieval for Multi-Hop Question Answering~\citep{Zhang2023EndtoEndBR} & NAACL 2024 \\
    \bottomrule
\end{tabular}}
\caption{\label{Paperlist}
    List of papers used in our Sci-Replicate benchmark.
}
\vspace{-1em}
\end{table}

\label{AnnotationProcess}
\paragraph{Step 1: paper selection}
We curated NLP papers from leading conferences in 2024, including ACL, EMNLP, ICLR, Neurips, and COLING.
Using a web crawler, we collected accepted paper titles and employed the PapersWithCode API\footnote{\url{https://paperswithcode.com/api/v1/docs/}} to identify those with open-source implementations.
For each identified paper, we retrieved corresponding GitHub repository links and metadata (e.g., stars, issues, release dates) via the GitHub REST API\footnote{\url{https://docs.github.com/en/rest?apiVersion=2022-11-28}}.

To filter candidates, we applied the following criteria:
\begin{itemize}
    \item Removed survey/exploratory papers while retaining method-focused research.
    \item Applied a cutoff date of January 1, 2024 to avoid data leakage.
    \item Excluded repositories with fewer than 5 stars to ensure basic quality assurance.
\end{itemize}
Subsequently, researchers manually reviewed each candidate paper and its repository. We discarded papers with excessive computational demands, poorly structured code, ambiguous documentation, missing preprocessing steps, or reported reproduction issues.

\paragraph{Step2: python environment setup}
For papers passing the initial screening, annotators followed the README to set up the environment and replicate experiments. Common issues included dependency conflicts, data loading failures, and incomplete or buggy code. Annotators attempted to resolve these problems; repositories with irrecoverable errors were excluded.
\paragraph{Step3: annotation}
Annotation consists of two steps:
\begin{enumerate}
    \item Algorithm-Function alignment: most papers contain multiple algorithmic components, often organized as subsections. Annotators segmented these into distinct units and mapped each to its corresponding implementation. Code was refactored to encapsulate each algorithm in a standalone function or method. Papers with implementations too fragmented for restructuring were excluded.
    \item Detailed annotation: for each aligned function, annotators documented input/output variables, intra- and cross-file dependencies, and external API usage. Additionally, they inserted explanatory comments mapping code segments to algorithm components. 
    Based on these annotations and variable dependencies, we can construct a reasoning graph representing the implementation logic.
    During the annotation process, LLMs were employed to assist with algorithm-function alignment and the generation of variable descriptions and code comments. All outputs were subsequently reviewed and corrected by human annotators to ensure accuracy.
\end{enumerate}

The final selected papers are listed in Table~\ref{Paperlist}.

\paragraph{Step 4: verification suite preparation} 
Finally, annotators created verification suites with 10 test cases per task, drawn from the original datasets used in each repository for the majority of papers.
For a small number of repositories, fewer than 10 test cases could be constructed. For instance, algorithms that analyze LLM parameters may have only a single test case.
Given the inherent randomness in many NLP implementations and potential machine-related variability, we addressed reproducibility from two angles:
\begin{itemize}
    \item Eliminating code randomness: annotators fixed random seeds and replaced non-deterministic operations (e.g., unordered sets) with deterministic equivalents to ensure consistent outputs across runs. 
    
    \item Controlling hardware variability: users were instructed to run both reference and generated code locally to eliminate discrepancies caused by hardware differences.
\end{itemize}
Lastly, annotators implemented task-specific comparison scripts to evaluate output correctness, accounting for variations in return types across tasks.

\section{Details of the Task Categories}
\label{Appen:taskcate}

\begin{figure}[h]
 	\centering
  \includegraphics[width=0.7\textwidth]{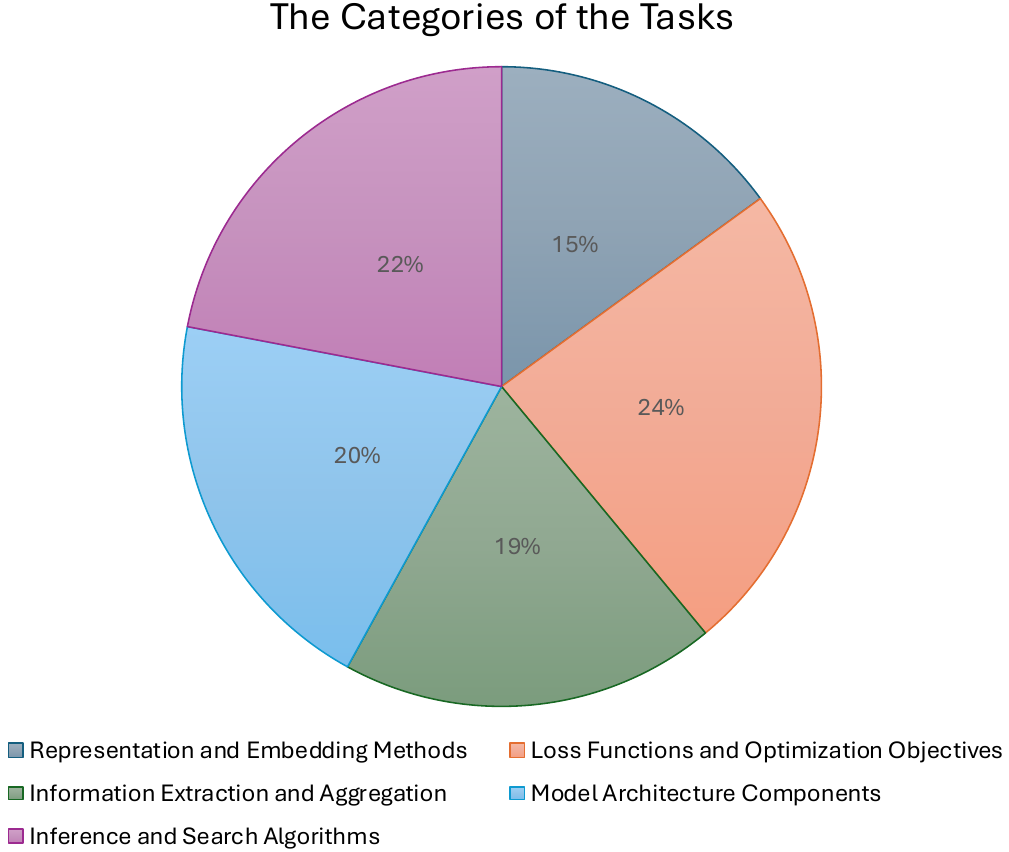} 
 	\caption{\label{taskcate}
 	The categories of the tasks within \DatasetName.}
 \end{figure}
 
 The benchmark encompasses five main task categories in the NLP domain: representation and embedding methods, loss functions and optimization objectives, information extraction and aggregation, model architecture components, and inference and search algorithms. 
 The distribution of each task category is illustrated in Figure~\ref{taskcate}.

\section{Details of the Actions}
\label{Appen:tool}
In this section, we provide implement details for all actions defined in the \ModelName.
\paragraph{SearchPaper}
We obtain the LaTeX source code of the target academic paper from arXiv\footnote{\url{https://arxiv.org/}} and apply regular expression-based parsing to extract the content corresponding to each section.
Subsequently, we iteratively feed the content of each subsection, along with the query generated by the large language model, into GPT-4o-mini. The model extracts relevant information and returns it as an observation to the paper agent.

\paragraph{SearchSection}
Following the same approach as \texttt{SearchPaper}, the tool begins by parsing the LaTeX source code of the target algorithm. Upon receiving a section ID from the Paper Agent, it retrieves and returns the content of the corresponding section.

\paragraph{SearchLiterature}
Given a paper ID and a query, the tool attempts to download the corresponding LaTeX source code from arXiv.
If the LaTeX source code is unavailable, it returns no information.
Otherwise, it extracts content relevant to the query from the paper, following the same procedure as the \texttt{SearchPaper} action.

\paragraph{SearchCode}
For each Python file in the code repository, we utilize the Python AST~\footnote{\url{https://docs.python.org/3/library/ast.html}} package to parse the file and extract all defined classes, functions, and global variables. 
Unlike embedding-based code search methods~\citep{Zhang2024SECONMS,Zhang2023I2RIA}, the Code Agent in our framework directly provides the name of a code item. The tool then returns the corresponding definition if it exists; otherwise, it returns an empty response.

\paragraph{SearchFile}
When the Code Agent provides a file name, the tool returns the full content of the corresponding file.
\paragraph{SearchWeb}
When the Code Agent issues a query, we use the Google Search API~\footnote{\url{https://developers.google.com/custom-search/v1/}} to retrieve relevant information from websites. 
These results are then processed by GPT-4o-mini, which filters the content and extracts the information most relevant to the query for return.
\paragraph{Compiler}
Once the Code Agent completes code generation, it invokes the compiler to execute the code. The generated function or method is inserted into the original Python file, and the corresponding Python environment is used to run the code. The output from the compiler is then returned as the feedback.

\section{Human Evaluation of Reasoning Graph Accuracy}

\begin{table}[h]
\centering
\caption{Reasoning Graph Accuracy of Sci-Reproducer Models Evaluated by LLMs and Human Annotators}
\label{tab:human_evaluation}
\begin{tabular}{lcc}
\toprule
\multirow{2}{*}{\textbf{Sci-Reproducer Model}} & \multicolumn{2}{c}{\textbf{Evaluation Method}} \\
\cmidrule(lr){2-3}
& \textbf{GPT-4o} & \textbf{Human (Mean ± Std)} \\
\midrule
GPT-4o-mini & 0.782 & 0.768 (0.066) \\
O3-mini-medium & 0.790 & 0.789 (0.033) \\
\bottomrule
\end{tabular}
\end{table}

\begin{table}[h]
\centering
\caption{Pearson Correlation between LLM and Human Judgments of Reasoning Graph Accuracy}
\label{tab:correlation}
\begin{tabular}{ccc}
\toprule
\textbf{Number of Data Points} & \textbf{r} & \textbf{P-value} \\
\midrule
20 & 0.7518 & $<$0.005 \\
\bottomrule
\end{tabular}
\end{table}

To validate the reliability of our LLM-based evaluation metrics, we conducted a human evaluation study on a subset of our benchmark. Three PhD students in computer science independently assessed the reasoning graph accuracy for 20 tasks generated by GPT-4o-mini and O3-mini-medium using Sci-Reproducer.

As shown in Table~\ref{tab:human_evaluation}, human evaluations demonstrate strong alignment with our automated LLM-based assessments. The mean scores show consistent agreement between human annotators and GPT-4o evaluations, with standard deviations indicating reasonable inter-annotator consistency.

Furthermore, we computed the Pearson correlation coefficient between human and LLM-based evaluations across all assessed instances. As presented in Table~\ref{tab:correlation}, the correlation (r = 0.7518, p < 0.005) indicates a strong positive relationship between human judgments and automated assessments, supporting the validity of our LLM-based evaluation approach.

While this preliminary validation demonstrates the effectiveness of our automated metrics, comprehensive human evaluation across the full benchmark remains an important direction for future work.

\section{Figures and Tables}
\label{FigureTable}

 \begin{figure}[h]
 	\centering
 \includegraphics[width=1\textwidth]{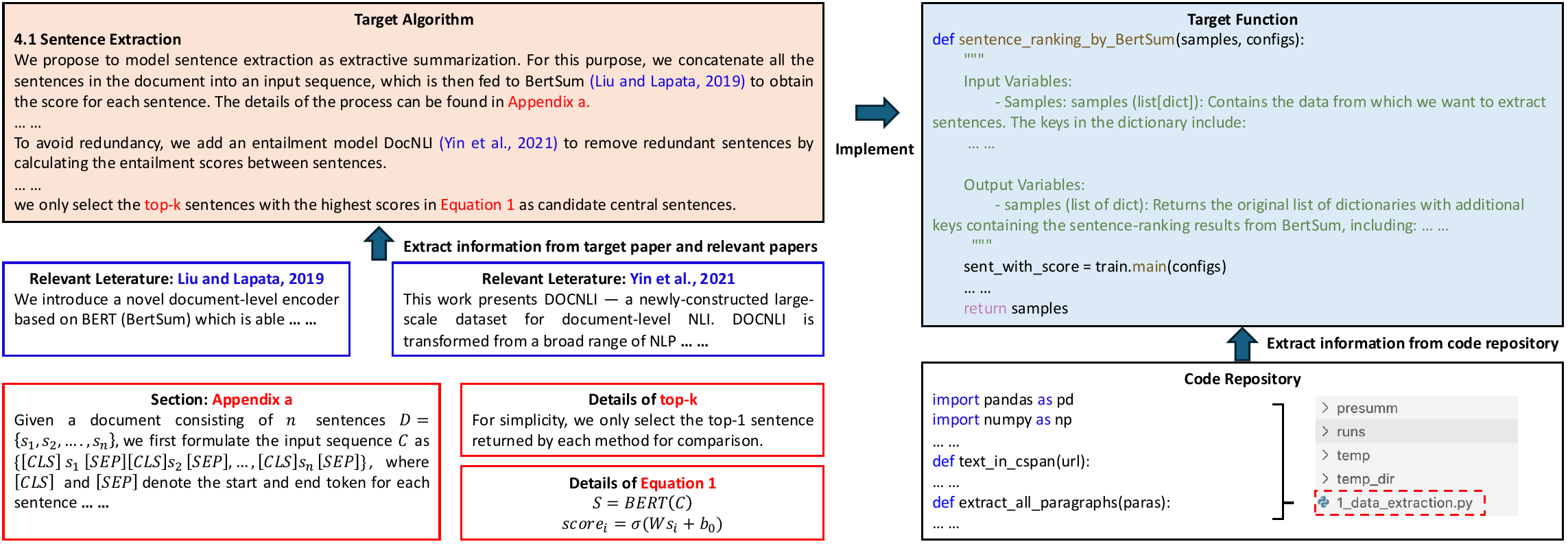} 
 	\caption{\label{taskintro}
 	The task consists of two steps: \textit{Algorithm Understanding} and \textit{Code Implementation}. (Left) The model must extract an algorithm’s workflow and details from the research paper, including descriptions and variable values from cited papers and other paper sections. (Right) Using this extracted information, the model implements the corresponding function in the code repository, correctly handling dependencies and API calls. }
 \end{figure}

\begin{figure}[h]
 	\centering
 \includegraphics[width=1\textwidth]{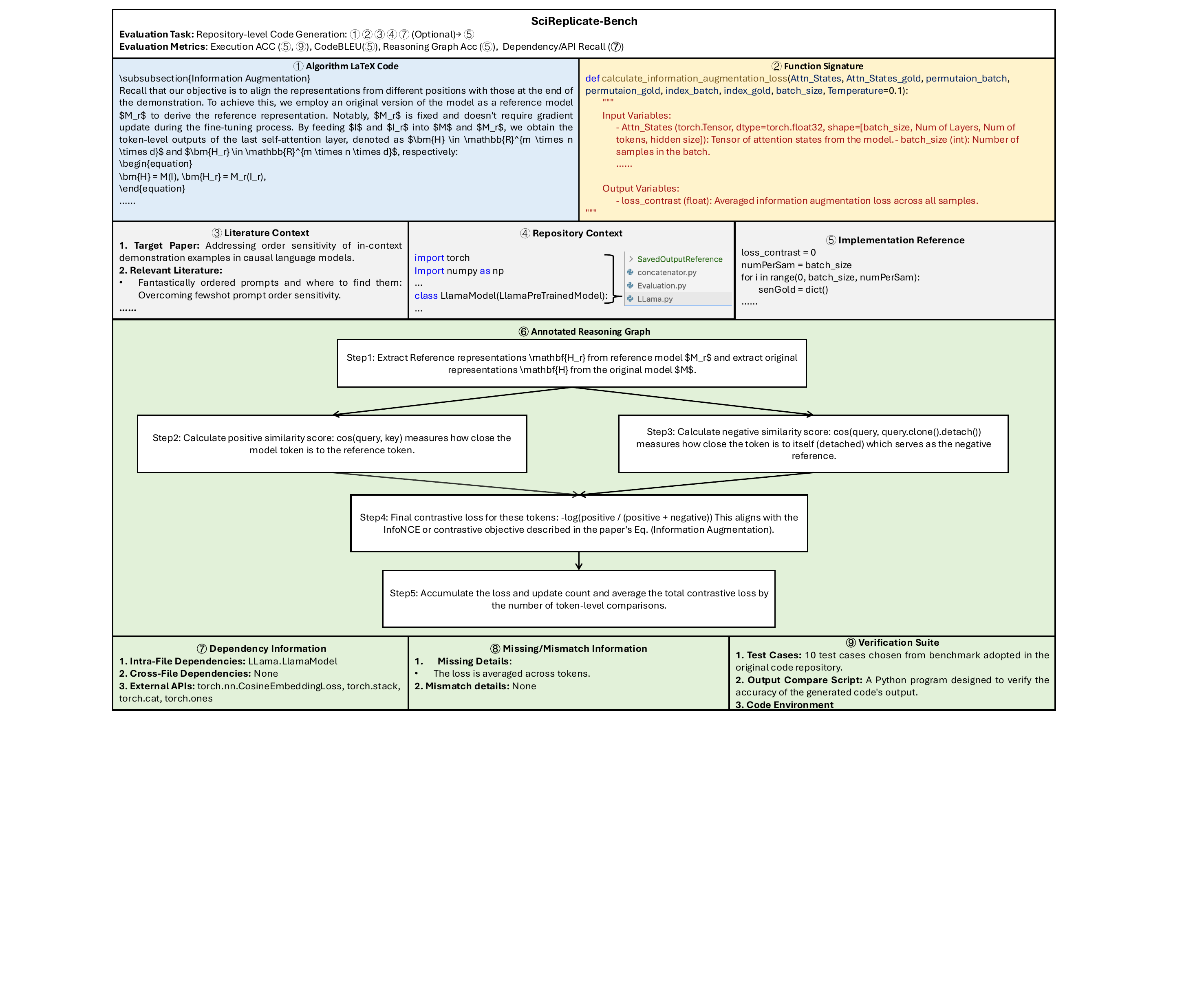} 
 	\caption{\label{Appen:datasetintro}
 	The overview of the \DatasetName.
    }
 \end{figure}

 \begin{table}[h]
\vspace{-1em} 
\centering
\resizebox{0.35\textwidth}{!}{  
\begin{tabular}{r|c}
    \toprule
    \textbf{Approach} & \textbf{Error Ratio} ($\downarrow$) \\
    \midrule
    No Agent & 80.3\\
    Paper Agent & 83.3 \\
    Code Agent & 29.4 \\
    \ModelName & \textbf{24.9} \\
    \bottomrule
\end{tabular}}
\caption{\label{Syntax}
    Syntax error across different settings.
}
\vspace{-1em}
\end{table}

 \begin{table}[h]
     \centering
     \resizebox{1\textwidth}{!}{
     \begin{tabular}{c|l}
     \toprule
Categories  & Examples \\
\midrule
\multirow{3}{*}{Hyperparameters and configurations}  & Thresholds, batch sizes, maximum iteration counts, exact numbers of clusters, \\
& initialization methods for variables or vectors, types of regularization~(such as L1 or L2) \\
& , and specific distance metrics (e.g., using L2 norm for Euclidean distances)\\
\midrule
\multirow{3}{*}{Numerical stability techniques} & clamping values to avoid numerical instability, adding small constants during   \\
& logarithmic calculations, managing division by zero scenarios, and addressing \\
& rounding and precision issues.\\
\midrule
\multirow{2}{*}{Implementation logic} &  Data splitting, application of dropout, formatting of input sequences, \\
& and handling special or edge cases in the input data.\\
\midrule
\multirow{2}{*}{Coding strategy} & Caching for performance enhancement,
retry mechanisms to handle failures, \\
& early stopping criteria, and strategies for memory
optimization.\\
\bottomrule
     \end{tabular}}
     \caption{Some examples for different missing information categories.}
     \label{tab:miss}
 \end{table}

\begin{figure}[h]
 	\centering
  \includegraphics[width=\textwidth, height=0.95\textheight, keepaspectratio]{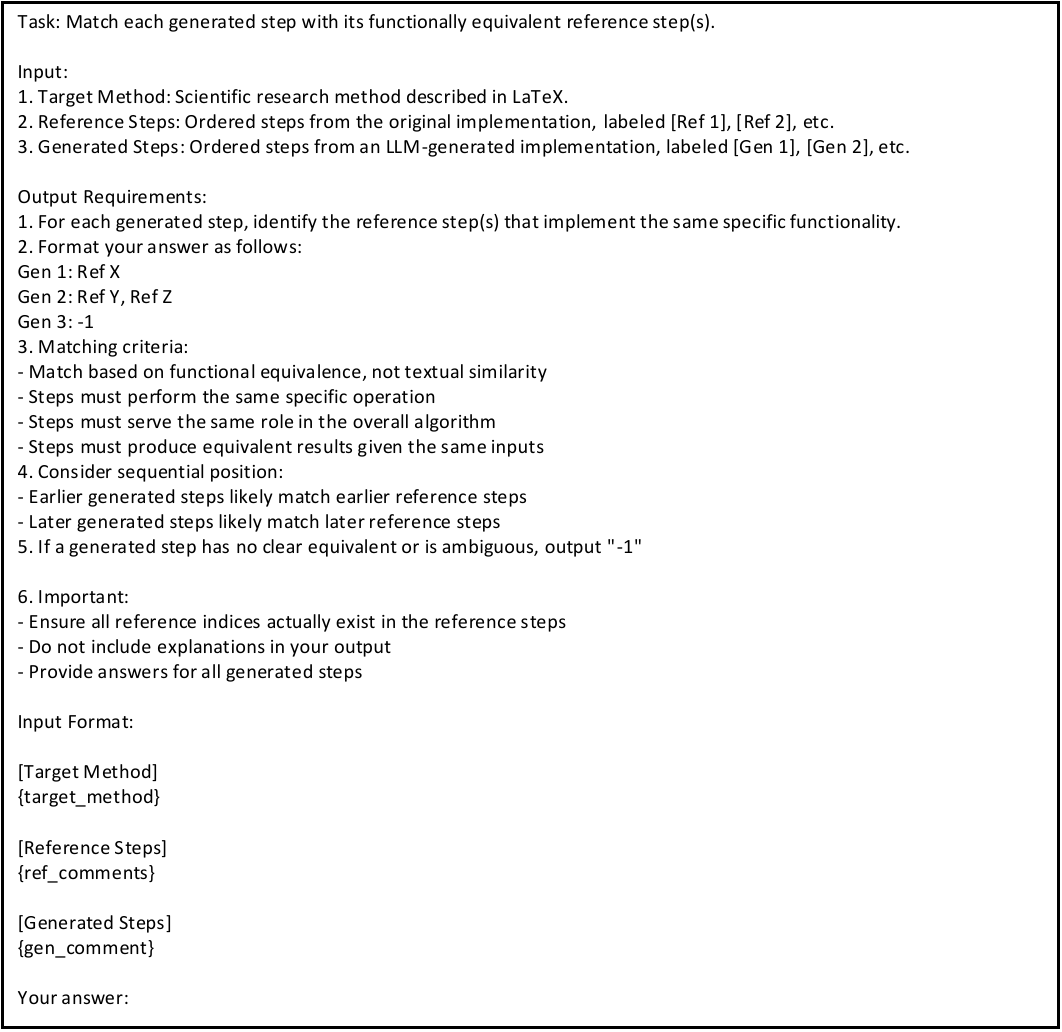} 
 	\caption{\label{PromptAlign}
 	The prompt for node matching.
    }
 \end{figure}

\begin{figure}[h]
 	\centering
  \includegraphics[width=\textwidth, height=0.95\textheight, keepaspectratio]{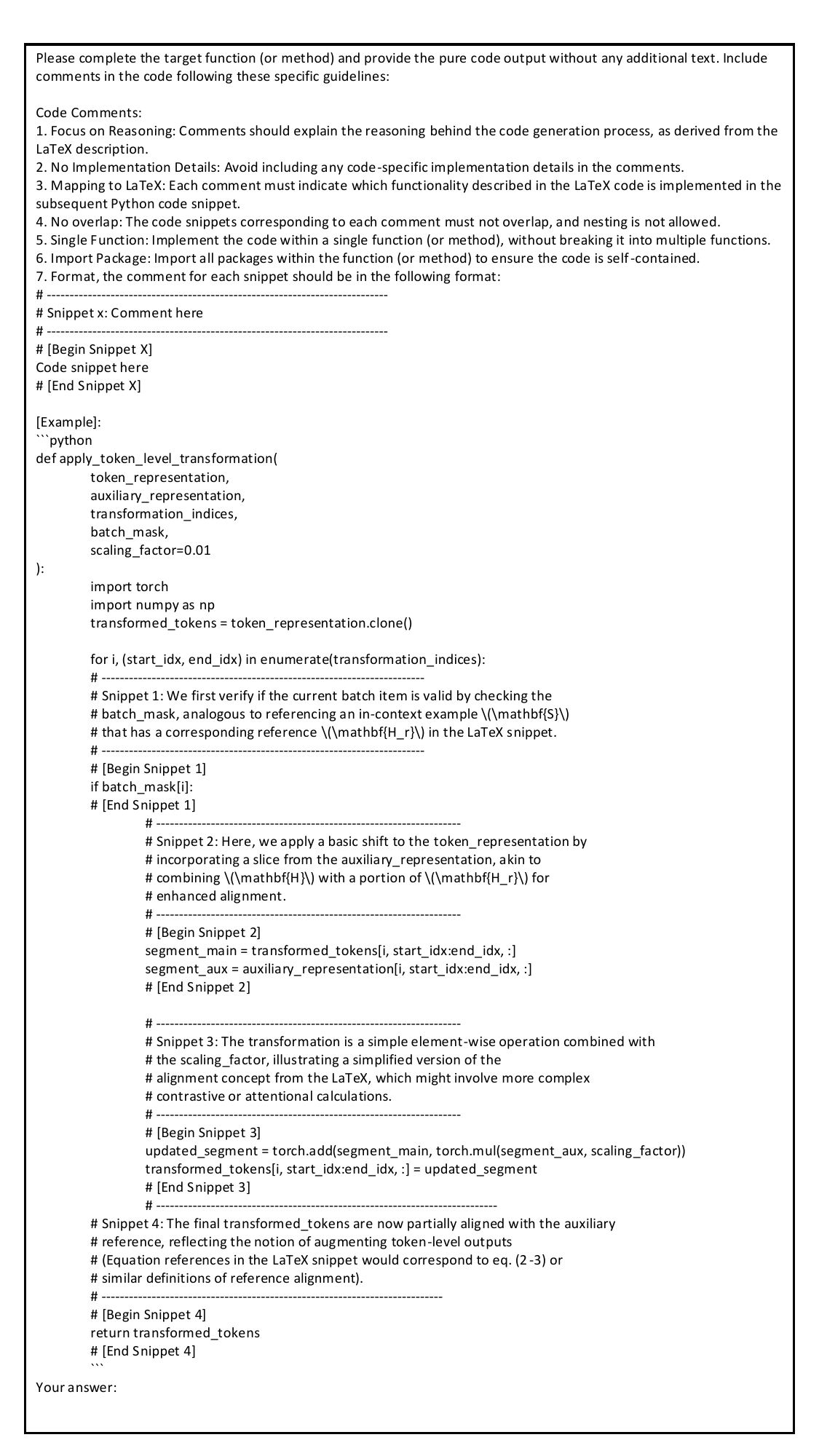} 
 	\caption{\label{PromptCodeGen}
 	The prompt for code generation.
    }
 \end{figure}

 \begin{figure}[h]
 	\centering
  \includegraphics[width=\textwidth, height=0.95\textheight, keepaspectratio]{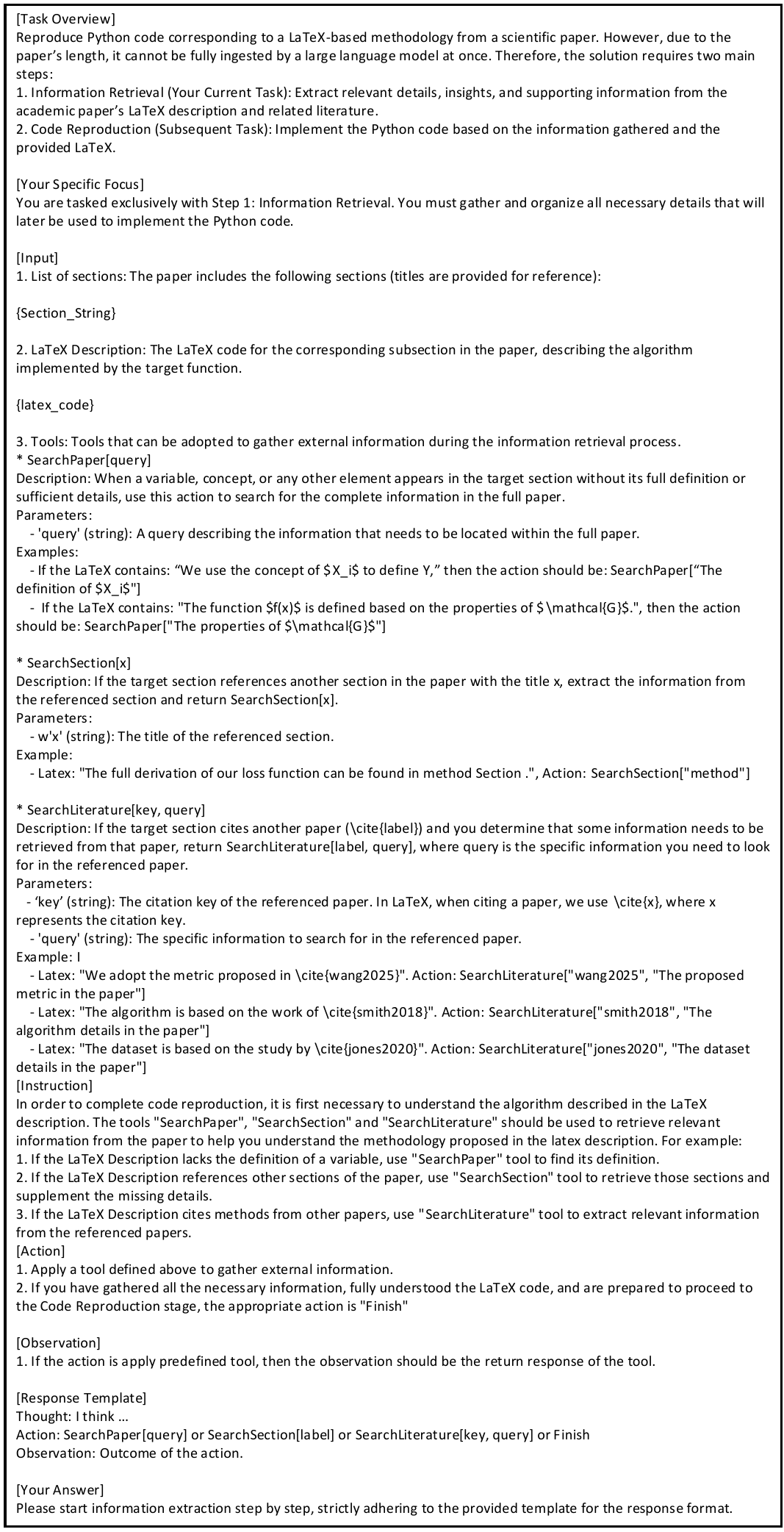} 
 	\caption{\label{PromptPaper}
 	The prompt for Paper Agent.
    }
 \end{figure}

\begin{figure}[h]
 	\centering
  \includegraphics[width=\textwidth, height=0.95\textheight, keepaspectratio]{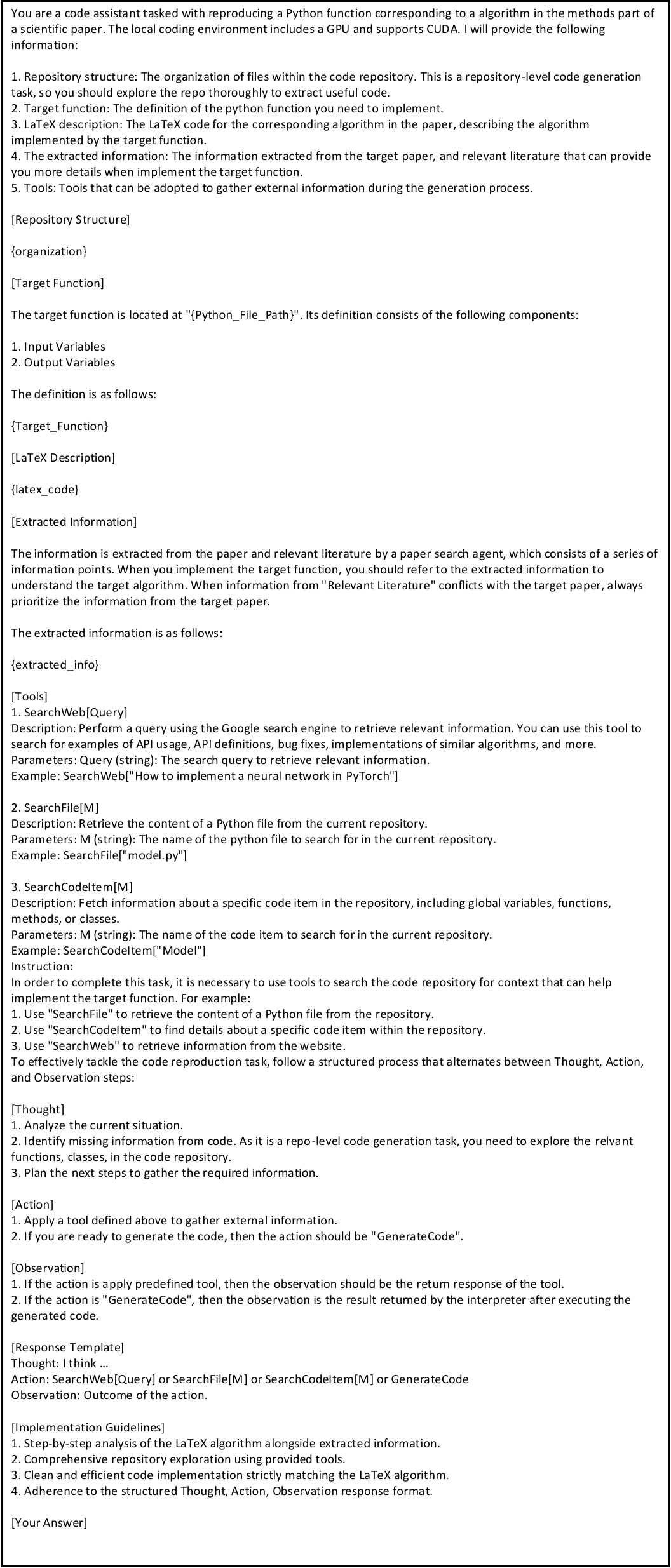} 
 	\caption{\label{PromptCode}
 	The prompt for Code Agent.
    }
 \end{figure}

 \begin{figure}[h]
 	\centering
  \includegraphics[width=\textwidth, keepaspectratio]{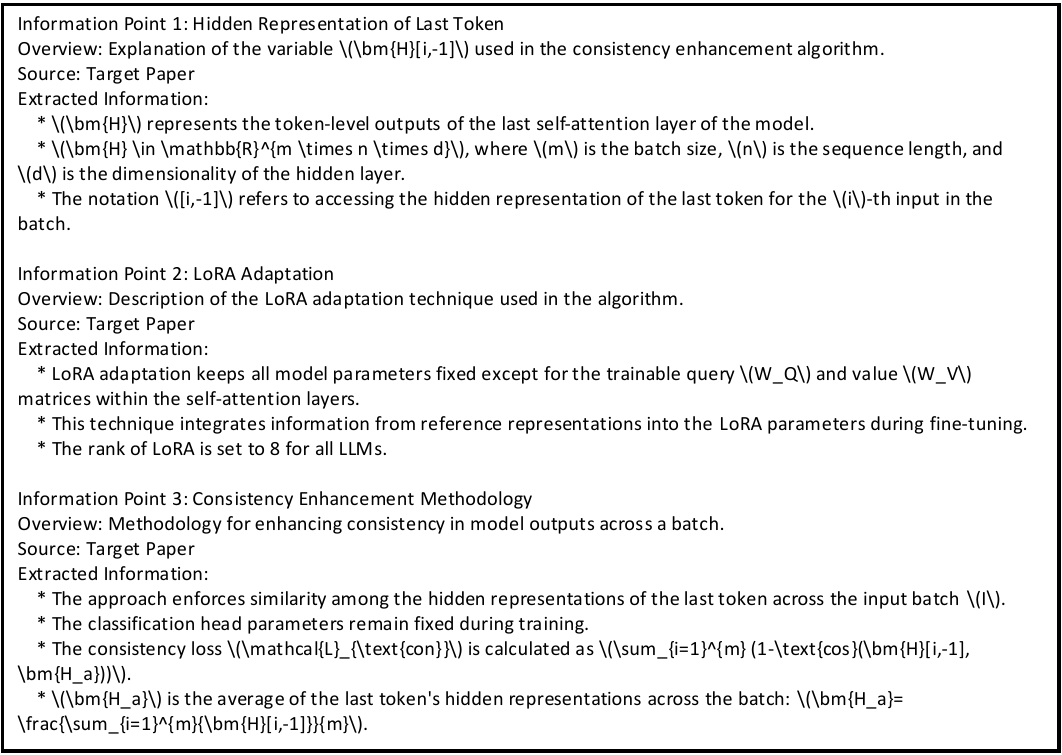} 
 	\caption{\label{PaperAgentOutput}
 	An example of output report of the Paper Agent.
    }
 \end{figure}

\end{document}